\documentclass[]{IEEEtran}

\usepackage{amsmath} 
\usepackage{amssymb}  
\usepackage{graphicx} 
\usepackage{subcaption} 
\usepackage{listings}
\usepackage{enumitem}
\usepackage[hyphens]{url}
\usepackage{hyperref}
\usepackage{wrapfig}
\usepackage{pgfplots}
\usepackage{afterpage}
\usepackage{multirow}

\DeclareMathOperator{\EX}{\mathbb{E}}
\pgfplotsset{compat=1.14}

\newcommand{\quotes}[1]{``#1''}

\title{Combining Experience Replay with Exploration by Random Network Distillation}

\author{Francesco Sovrano
	\thanks{The authors are at the University of Bologna, Department of Computer Science and Engineering (DISI),
		Mura Anteo Zamboni 7, 40127, Bologna, Italy}}

\begin{document}
	
	\IEEEpubid{\begin{minipage}{\textwidth}\ \\[12pt]
			978-1-7281-1884-0/19/\$31.00 \copyright 2019 IEEE
	\end{minipage}}
	\maketitle
	
	\begin{abstract}
		Our work is a simple extension of the paper \quotes{Exploration by Random Network Distillation}\cite{burda2018exploration}. More in detail, we show how to efficiently combine Intrinsic Rewards with Experience Replay in order to achieve more efficient and robust exploration (with respect to PPO/RND) and consequently better results in terms of agent performances and sample efficiency. We are able to do it by using a new technique named Prioritized Oversampled Experience Replay (POER), that has been built upon the definition of \textit{what is the important experience useful to replay}. Finally, we evaluate our technique on the famous Atari game \textit{Montezuma's Revenge} and some other hard exploration Atari games.
	\end{abstract}
	
	\begin{IEEEkeywords}
		Deep Reinforcement Learning, Actor-Critic, Prioritized Experience Replay, PPO, Intrinsic Rewards, Montezuma's Revenge
	\end{IEEEkeywords}

	\section{INTRODUCTION}
	\IEEEPARstart{A}{ Reinforcement Learning} (RL) problem is typically formalized as a Markov Decision Process (MDP), in which an agent interacts with an environment, observing the effects of its actions (in the environment) while trying to maximize a cumulative return/reward. In other words, a RL agent learns how to optimally interact with the environment, by receiving some environmental feedbacks called rewards. The more an action is good, the higher should be the reward. But in many scenarios, rewards are very rare and difficult to get, thus making Reinforcement Learning very hard to achieve. \\
	A good RL agent has to be able to efficiently explore the environment while optimally exploiting the information it already found. Balancing exploration with exploitation is a well-known problem in RL, very hard to solve when the environment is too big and the rewards are too sparse and difficult to find by taking random sequences of actions (eg. in the Atari game \textit{Montezuma's Revenge}). \\
	In literature, several techniques exist to improve the sample efficiency (the ability to exploit past information) or the exploratory skills of existing RL algorithms. \\
	For example, many techniques used to improve sample efficiency are based on Experience Replay (ER). ER consists in storing, into a buffer, information (experience) about the agent actions in the environment and then replaying this experience during training. The key idea behind ER is to store memory of important and meaningful states in order to exploit it, but replaying such memory can introduce some bias. In other words, the agent might tend not to explore new states, because it is too focused on exploiting the old information. \\
	On the other side, many techniques for improving exploration exist in the RL literature. Among them we cite the \quotes{exploration by random network distillation}\cite{burda2018exploration} (PPO/RND). PPO/RND is a recent breakthrough in Actor-Critic based RL, because it has given a significant progress on several hard exploration problems, such as the famous Atari game \textit{Montezuma's Revenge}. PPO/RND uses Proximal Policy Optimization (PPO) \cite{schulman2017proximal} in conjunction with Random Network Distillation (RND). RND is a recent technique for Intrinsic Motivation based on prediction errors, in which the environmental feedback (the reward) is augmented with an extra intrinsic reward that is proportional to the error of a neural network predicting features of the observations given by a fixed randomly initialized neural network \cite{burda2018exploration}. \\
	Our work is a simple extension of PPO/RND. We show how to efficiently combine Intrinsic Rewards with Experience Replay in order to achieve more efficient and robust exploration than PPO/RND and consequently better results in terms of agent performances and sample efficiency. We are able to do it by using a new technique named Prioritized Oversampled Experience Replay (POER), that has been built upon the definition of \textit{what is the important experience useful to replay}. In POER we mix oversampling \cite{jaderberg2016reinforcement} with experience prioritization \cite{schaul2015prioritized}, trying to achieve \textit{the goal of an optimal balance between exploration and exploitation}. In order to do this, we:
	\begin{itemize}
		\item give a definition of \textit{important} information
		\item find a way to know when information is \textit{uncommon}
		\item understand how to use important uncommon information to improve the exploratory skills of the agent
	\end{itemize}
	
	More in detail, with our experiments we show how POER affects the average cumulative return of a baseline PPO/RND agent in the following hard exploration \cite{bellemare2016unifying} Atari games: \textit{Montezuma's Revenge, Solaris, Venture}. \\
	Interestingly we find that our technique seems to properly balance exploration and exploitation especially in \textit{Montezuma's Revenge}, while in some other games it does not \footnote{maybe due to the insufficient amount of training time, or due to the adopted replay frequency}.
	
	\subsection{Structure of the article}
	In section \ref{sec:related_work} we introduce some related works, providing at section \ref{sec:background} the necessary background information about Reinforcement Learning. In section \ref{sec:combining_IR_with_ER} we show how to combine Experience Replay and Intrinsic Rewards in PPO. While in section \ref{sec:experiments} we describe the results of our experiments through an ablative analysis, trying to highlight the complexity behind combining experience replay and intrinsic rewards, by showing among other things:
	\begin{itemize}
		\item How a too frequent experience replay can negatively affect the agent performances.
		\item How an accurate choice of what kind of experience to replay can impact on the agent performances.
	\end{itemize}
	
	\section{Related work} \label{sec:related_work}
	Our work is an extension of PPO/RND\cite{burda2018exploration}. PPO/RND is a recent breakthrough in Actor-Critic based Reinforcement Learning, because it has given a significant progress on several hard exploration problems, such as the famous Atari game \textit{Montezuma's Revenge}. PPO/RND uses Proximal Policy Optimization (PPO) \cite{schulman2017proximal} in conjunction with Random Network Distillation (RND). RND is a recent technique for Intrinsic Motivation based on prediction errors, in which the intrinsic reward is the error of a neural network predicting features of the observations given by a fixed randomly initialized neural network \cite{burda2018exploration}. Furthermore, PPO/RND introduces a simple method to flexibly combine intrinsic and extrinsic rewards. \\
	Our work extends PPO/RND\cite{burda2018exploration} with a new Prioritized Oversampled Experience Replay technique. The goal of our work is to combine the sample efficiency provided by experience replay with the exploratory properties of RND, in order to achieve more efficient and robust exploration and consequently better results in terms of agent performances and sample efficiency. \\
	There have been several (successful) attempts to combine Actor-Critic algorithms (eg. PPO) with Experience Replay. Probably some of the most interesting are:
	\begin{itemize}
		\item \quotes{ACER}\cite{wang2016sample}, a sophisticated technique based on trust region policy optimization
		\item \quotes{Self-imitation learning}\cite{oh2018self}, a simple prioritized experience replay technique applied to PPO
	\end{itemize} 
	The main differences between our work and \cite{oh2018self} are that the latter:
	\begin{itemize}
		\item does not make any use of intrinsic rewards
		\item does not use any oversampling technique integrated in the experience replay mechanism
	\end{itemize}
	Furthermore \quotes{Self-imitation learning}\cite{oh2018self} prioritizes experience using the \textit{Advantage} function while our technique prioritizes experience using \textit{Intrinsic Rewards}. In section \ref{sec:experiments} we show a comparison of the aforementioned prioritization approaches. \\
	Anyway, trying to combine exploration with exploitation is historically a challenging problem in modern RL. Among all the works related to this problem we cite \cite{yang2019never}: a new and interesting approach that tries to balance between exploration and exploitation by employing optical flow estimation errors to examine the novelty of new observations and deliver permanent performance without encountering catastrophic forgetting problems.
	
	\section{Reinforcement Learning Background} \label{sec:background}
\begin{figure}[t]
		\centering
		\includegraphics[width=.48\textwidth]{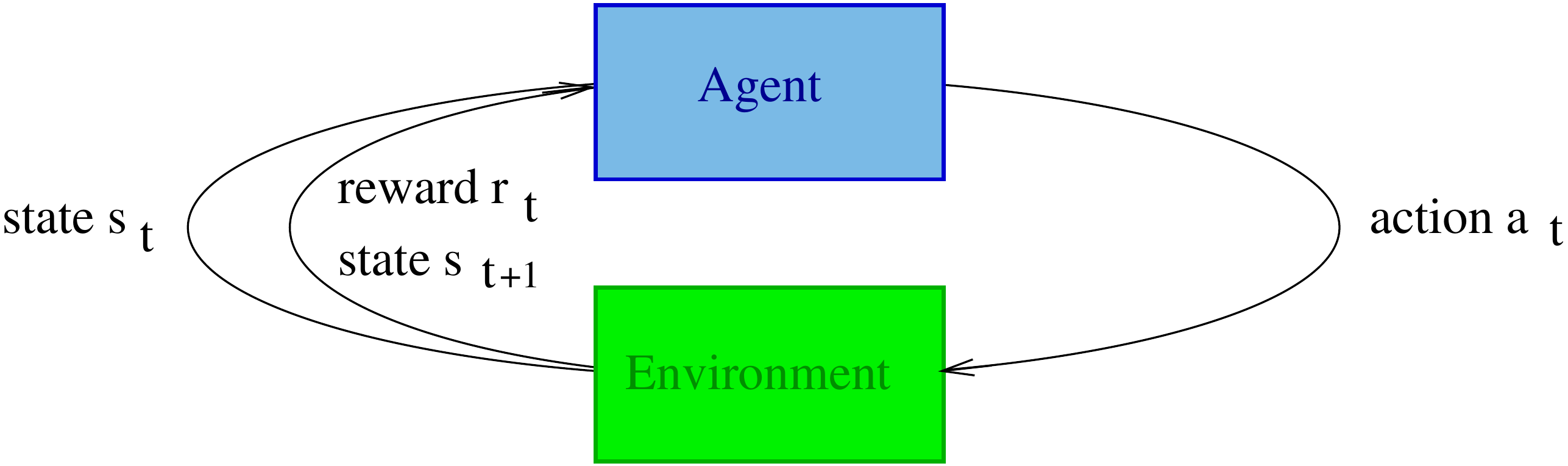}
		\caption{Basic operations of a Markov Decision Process}
		\label{fig:MDP}
	\end{figure}

\begin{figure}[ht]
		\centering
		\includegraphics[width=.5\textwidth]{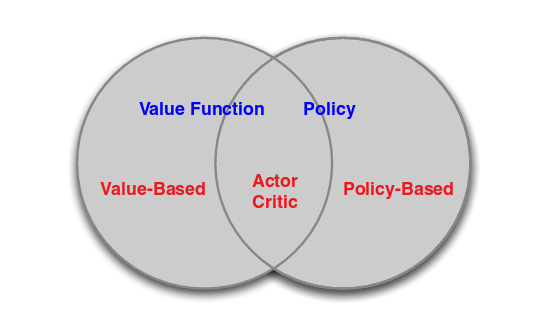}
		\caption{\textbf{Reinforcement Learning}: Algorithms families}
		\label{fig:families}
	\end{figure}

	This section contains a short introduction to Reinforcement Learning techniques, mostly with the aim to fix notation. The content is quite standard, and we largely borrowed it from our previous works \cite{amslaurea16718,asperti2019crawling,asperti2018crawling}. \\
	A Reinforcement  Learning  problem is typically formalized as a Markov Decision Process (MDP). In this setting, an agent interacts at discrete time steps with an external environment. At each time step $t$,  the agent observes a state $s_t$ and chooses an action $a_t$ according to some policy $\pi$, that is a mapping (a probability distribution) from states to actions. As a result of its action, the agent obtains a reward $r_t$ (see Fig.~\ref{fig:MDP}), and the environment passes to a new state $s'=s_{t+1}$.
	The process is then iterated until a terminal state is reached.
	The future cumulative reward $R_t = \sum^\infty_{k=0}\gamma^{k}r_{t+k}$ is the total cumulated reward from time starting at $t$. $\gamma \in \left[ 0, 1 \right]$ is the so called {\em discount factor}: it represents the difference in importance between present and future rewards. \\
	The goal of the agent is to maximize the expected cumulative return starting from an initial state $s=s_t$. \\
	The  {\em action  value} $Q^\pi(s,a)  = \EX^\pi[R_t|s=s_t,a=a_t]$ is the expected return for selecting action $a$ in state $s_t$ and prosecuting with strategy $\pi$. \\
	Given a state $s$ and an action $a$, the optimal {\em action value} function $Q^*(s,a) = \max_\pi Q^\pi(s,a)$ is the best possible action value achievable by any policy. \\
	Similarly, the {\em value} of state $s$ given a policy $\pi$ is $V^\pi(s)  = \EX^\pi[R_t|s=s_t]$ and the optimal value function is $V^*(s) = \max_\pi V^\pi(s)$. \\
	The starting points for the RL methodology are two fundamental dynamic programming algorithms: value iteration and policy iteration. In the value-based approach, we define the parameters of a value function that quantifies the maximum cumulative reward obtainable from a state belonging to the state space. While, in the policy-based approach, the policy parameters are tuned in a direction of improvement. \\ In figure \ref{fig:families} a simple overview of the RL families is shown.
	Value-based approaches (eg. DQN\cite{mnih2013playing}) try to find a policy that maximizes the cumulative return by keeping a set of expected returns estimates for some policy $\pi$. Usually $\pi$ is either the \quotes{current} policy or the optimal one. An algorithm that uses the current policy is called \textit{on-policy} algorithm, while an algorithm using the optimal one is called \textit{off-policy}. \\ 
	In policy-based RL, instead of parametrizing the value function and doing $\epsilon$-greedy policy improvements to the value function parameters, we parametrize the policy $\pi_\theta(a|s)$ and update the parameters $\theta$ descending gradient into a direction that improves $\pi$, because sometimes the policy is easier to approximate than the value function. \\ 
	Actor-Critic methods are in the middle between policy-based and N-step value-based RL. In Actor-Critic we have two function approximators: one for the policy (the Actor) and one for the value function (the Critic).
	
	\subsection{Advantage Actor-Critic and PPO} \label{sec:ActorCritic}
	The idea behind Actor-Critic is that it is possible to reduce the variance of the policy while keeping it unbiased by subtracting a learned state-value function $V(s)$ known as baseline. $R - V(s)$ is called the advantage function $A(s)$. The advantage is used to measure how much better than \quotes{average} it is to take an action given a state. \\
	Thus, the Actor updates its parameters $\theta_1$ in the direction of:
	\begin{equation} \label{eq:loss_ac_policy}
	\EX [A(s) \cdot \nabla_{\theta_1} \log \pi_{\theta_1}(a | s)]
	\end{equation}
	While the Critic updates its parameters $\theta_2$ in the direction of:
	\begin{equation} \label{eq:loss_ac_value}
	c_1 \nabla_{\theta_2} \Omega \left( R - V^{\theta_2}(s) \right)
	\end{equation}
	Where $\Omega$ usually is the L2 loss \cite{L2Loss}, and $c_1$ is a regularization constant used to keep the Critic learning rate lower than the Actor. Commonly $c_1 < 1$ (eg: $c_1 = \frac{1}{2}$). But it is not uncommon to have $c_1 = 1$. \\
	The Actor-Critic family can count many different algorithms including: A3C\cite{mnih2016asynchronous}, A2C\cite{schulman2017proximal}, PPO\cite{schulman2017proximal}. \\

	\subsubsection{\textbf{A3C}} \label{sec:a3c}
	The Asynchronous Advantage Actor-Critic (A3C) algorithm \cite{mnih2016asynchronous} is an on-policy technique based on Actor-Critic. A3C, instead of using an experience replay buffer like DQN\cite{mnih2013playing}, uses multiple agents on different threads to explore the state spaces, and it makes decorrelated updates to the Actor and the Critic. 
	It is important to mention that A3C uses a different version of the policy gradient formula in order to better tackle the exploration problem. Thus, in A3C, the Actor objective function is changed to:
	\begin{equation} \label{eq:a3c_policy_loss}
	J^{\text{ON}}({\theta_1}) = \EX \left[ A(s) \cdot \log \pi_{\theta_1} (a|s) - \beta S(\pi_{\theta_1} | s) \right]
	\end{equation}
	Where $S(\pi_\theta | s)$ is the entropy of policy $\pi_\theta$ given state $s$, and $\beta$ is an entropy regularization constant.
	
	\subsubsection{\textbf{A2C}} \label{sec:a2c}
	In A3C each agent talks to the global parameters independently, so it is possible sometimes the parallel agents would be playing with different policies and therefore the aggregated update would not be optimal. The aim of A2C is to resolve this inconsistency. 
	A2C uses a coordinator that waits for all the parallel actors to finish before updating the global parameters, for this reason A2C is said to be the synchronous version of A3C. \cite{stooke2018accelerated}
		
	\subsubsection{\textbf{PPO}}
	Proximal Policy Optimization (PPO) \cite{schulman2017proximal} is an Actor-Critic algorithm based on A2C. \\
	The idea behind PPO is that, in order to improve training stability, we should avoid parameter updates that change the policy too much at one step. PPO is a simpler variation of Trust Region Policy Optimization  (TRPO) \cite{schulman2015trust} that prevents (too) big changes to the policy parameters by clipping them in a predefined range. \\
	PPO is much simpler to implement than TRPO, more general, and according to \cite{schulman2017proximal} it has empirically better sample complexity. \\
	In order to understand PPO, first of all lets denote the \textit{probability ratio} between the old policy and the new one as:
	\begin{equation}
	r(\theta_1) = \frac{\pi_{\theta_1}(a|s)}{\pi_{{\theta_1}^{\text{old}}}(a|s)}
	\end{equation}
	Then, the new Actor's objective function for PPO is:
	\begin{equation}
	\EX \left[ A(s) \cdot r(\theta_1) - \beta S(\pi_{\theta_1} | s) \right]
	\end{equation}
	Let $\hat{r}(\theta_1) = \text{clip} \left( r(\theta_1),1-\epsilon, 1+\epsilon \right)$, then the Actor objective function of PPO is:
	\begin{equation} \label{eq:PPO_Actor_loss}
	J^{\text{PPO}}(\theta_1) = \EX \left[ \min \left( r(\theta_1) \cdot A(s), \hat{r}(\theta_1) \cdot A(s) \right) - \beta S(\pi_{\theta_1} | s) \right]
	\end{equation}
	where $\epsilon$ is the clipping range hyper-parameter. \\
	In PPO the Critic uses the same clipping technique used by the Actor, but instead of keeping the minimum between the clipped and the non-clipped objective, it keeps the maximum. 
	Let $\hat{V}^{\theta_2}(s) = V^{{\theta_2}^\text{old}}(s) + \text{clip} \left( V^{\theta_2}(s) - V^{{\theta_2}^\text{old}}(s),-\epsilon, \epsilon \right)$, the objective function of the Critic is:
	\begin{equation}
	J^{\text{PVO}}(\theta_2) = c_1 \max \left( \Omega(R - V^{\theta_2}(s)), \Omega(R - \hat{V}^{\theta_2}(s)) \right)
	\end{equation}
	with $c_1 = 0.5$.
	
	\subsection{Experience Replay} \label{sec:exp_replay}
	Experience Replay \cite{lin1992memory} is actually a valuable and common tool for RL that has gained popularity thanks to Deep Q-learning \cite{Qlearning15}. \\
	The benefits coming from experience replay are:
	\begin{itemize}
		\item More efficient use of previous experience
		\item Less sample correlation, giving better convergence behaviour when training a function approximator
	\end{itemize}
	\textit{Importance sampling} is probably one of the most used techniques to implement efficient experience replay mechanisms in Actor-Critic algorithms (eg. ACER \cite{wang2016sample}). The idea behind \textit{importance sampling} is to weight the action gain of an old policy according to its relevance with respect to the current policy. \\
	Furthermore, to improve learning performances it is possible to prioritize experience (instead of sampling it uniformly) in order to replay important experience more frequently \cite{schaul2015prioritized}. \\
	It is interesting to note that the \textit{probability ratio} adopted in the PPO loss (eq. \ref{eq:PPO_Actor_loss}) is exactly the same importance weight used in importance sampling. This probably makes PPO already suitable for efficient experience replay.
	
	\subsection{Intrinsic Rewards}
	Improving exploration may be a complex task and it can be achieved in several ways. Some of them requires changing the gradient formula in order to maximize policy entropy (as in A3C), but usually entropy regularization is not enough. Another interesting approach consists in giving \textit{intrinsic rewards}, in order to motivate exploration/curiosity. Intrinsic rewards are rewards that do not involve receiving feedback from the environment (the \quotes{outside}), in fact they are a feedback from the agent itself (the \quotes{inside}) and for this reason they completely differ from usual rewards in RL (the extrinsic rewards, from \quotes{outside}). \\
	Most formulations of intrinsic rewards for exploration can be grouped into two broad classes \cite{bellemare2016unifying,pathak2017curiosity}:
	\begin{itemize}
		\item Count-Based \cite{ostrovski2017count,tang2017exploration}: the intrinsic reward is inversely proportional to the number of times a new state has been seen.
		\item Intrinsic Motivation \cite{burda2018exploration,chentanez2005intrinsically,schmidhuber1991possibility,still2012information,csimcsek2006intrinsic}: the intrinsic reward is proportional to how much the new state contains new information.
	\end{itemize}
	Count-Based methods in practice tends to fail with huge state spaces, while Intrinsic Motivation is usually more difficult to achieve and generalize. An interesting attempt to unify Count-Based methods and Intrinsic Motivation has been shown in \cite{bellemare2016unifying}. \\
	Among the Intrinsic Motivation techniques for exploration we cite Random Network Distillation \cite{burda2018exploration} (RND): a recent technique based on prediction errors, in which the intrinsic reward is the error of a neural network (the Predictor) predicting features of the observations given by another fixed randomly initialized neural network (the Target). The Predictor is trained to minimize its predictions error. 
	
	\section{Combining Experience Replay and Intrinsic Rewards, in PPO} \label{sec:combining_IR_with_ER}
	PPO\cite{schulman2017proximal} is probably one of the best state-of-the-art actor-critic algorithms, also because of its simplicity and versatility. As briefly shown in section \ref{sec:ActorCritic}, PPO already uses importance weights in its loss and this makes PPO already suitable for efficient experience replay without any need to re-integrate importance sampling. \footnote{this is more probably the reason behind the results of \quotes{Self-imitation learning}\cite{oh2018self}} \\
	The goal of our paper is to show how to combine \quotes{Experience Replay} and \quotes{Intrinsic Rewards}, two different and conflicting techniques, within PPO.
	More in detail: in this section we show how to combine PPO/RND\cite{burda2018exploration} with a new Experience Replay (ER) mechanism inspired by \cite{jaderberg2016reinforcement} and \cite{schaul2015prioritized}. 
	PPO/RND is a recent breakthrough in Actor-Critic RL, because it has given a significant progress on several hard exploration Atari games. PPO/RND uses PPO in conjunction with RND. PPO/RND introduces a simple method to flexibly combine intrinsic and extrinsic rewards by using two separate Critics for intrinsic and extrinsic \textit{values}, and then mixing these \textit{values} together in the final \textit{advantage} by a weighted sum that gives more importance to the extrinsic advantage.
	
	\subsection{Exploration vs Exploitation}
	Exploration and exploitation are usually seen as two opposite sides of the same coin. It is hard to optimally balance them. \\
	Exploration means: 
	\begin{itemize}
		\item explore the state space
		\item gather more information
	\end{itemize}
	Exploitation means: 
	\begin{itemize}
		\item exploit the already seen state space
		\item make the best decision given current information
	\end{itemize}
	Balancing exploration with exploitation may be a very important task to accomplish, because in many realistic scenarios the best long-term strategy may involve short-term sacrifices, and gathering enough information to make the best overall decisions is usually not a naive process. \\
	Lets try to understand it with an example. Lets assume that:
	\begin{itemize}
		\item we are playing a very complex turn-based strategy game with a partially observable environment
		\item we can perform only one move per turn
	\end{itemize}
	For simplicity, lets say that in this game there are two possible strategies we can exclusively follow:
	\begin{itemize}
		\item We can play the move we believe is best: this is called Exploitation
		\item We can play an experimental move: this is called Exploration
	\end{itemize}
	As you may intuitively imagine, optimally choosing between these two strategies might not be trivial at all. In RL we can say the same about combining Experience Replay (for exploitation) with Intrinsic Rewards (for exploration). For example, in section \ref{sec:experiments} we will show how a too frequent experience replay can negatively affect the agent performances.
	
	\subsection{Experience Replay Prioritization and Oversampling} \label{sec:er_implementation}
	ER consists in storing, into a \textit{buffer}, information (experience) about the agent actions in the environment and then replaying this experience during training. The experience is stored into mini-batches containing information about performed actions, seen states, rewards, etc.. Every mini-batch contains information about $B_s$ consecutive steps of an episode, where $B_s$ is called batch size. The key idea behind ER is to store memory of \textbf{important and uncommon batches} in order to exploit it. But replaying this memory can introduce some bias, worsening the exploratory skills of the agent. \\
	In order to properly use ER within PPO we need to:
	\begin{itemize}
		\item give a definition of \textit{important} batch
		\item find a way to know when a batch is \textit{uncommon}
		\item understand how to use important uncommon batches to improve the exploratory skills of the agent
		\item understand how to efficiently implement ER
	\end{itemize}

	\subsubsection{Definition of important batch}
	We say that a batch is important if it belongs to one of the following \textbf{importance classes}:
	\begin{itemize}
		\item The class of batches that contain positive extrinsic rewards.
 		\item The class of batches that lead to positive extrinsic rewards (without containing such rewards).
 		\item The class of batches that may probably lead to unseen states and consequently to new positive extrinsic rewards.
	\end{itemize}
	Usually the first and the second importance classes are under-represented in hard exploration games/problems (eg. some famous Atari games: \textit{Montezuma's Revenge}, \textit{Freeway}, etc.. \cite{bellemare2016unifying}). Thus, taking inspiration from \cite{jaderberg2016reinforcement}, we decided to \textbf{oversample} the important batches, by building $3$ different experience buffers (one for each class) and by uniformly sampling from them (during replay).
	
	\subsubsection{How to identify and use uncommon batches}
	Until now, we have defined what important batches are, but we still have not defined how to know when a batch is uncommon. We say that a batch is \textbf{uncommon} when it has a \textbf{high cumulative intrinsic reward}. Thus, we can detect uncommon batches through RND. In fact, when the cumulative intrinsic reward (generated by RND) of a batch is high, then we can say that the states contained in the batch are likely to be uncommon. \\
	We can exploit this information by \textbf{prioritizing} experience replay with the cumulative intrinsic rewards of the batches, in order to mainly replay important and uncommon batches. The result is a new Prioritized Oversampled Experience Replay technique meant to work in conjunction with intrinsic rewards and able to improve the efficiency of RND in terms of state-space exploration. 
	
	\subsubsection{ER implementation} \label{sub:er_implementation}
	We know what important batches are and how to prioritize them in order to improve exploration efficiency, but we still have not properly defined how we can concretely implement an efficient Experience Replay (ER) mechanism. A (too) naive Prioritized ER algorithm may add too much complexity to the learning algorithm making experience replay impracticable. For this reason, we decided to use a simple variation of the efficient algorithm proposed in \cite{schaul2015prioritized}. More in detail, our ER algorithm uses a circular buffer called Experience Buffer. This buffer has a fixed size $B$ and it is fed with new batches until it is completely full. When the buffer is full we have to \textbf{drop an old batch} in order to insert a new one, thus we compute a random number $p_d$ in $[0,1]$ and if $p_d$ is lower than the drop probability $P_d$, then the batch having the lowest priority is replaced, otherwise the batch at position $i \mod B$ is replaced and $i$ is incremented by $1$. \\
	The replay frequency is defined by a Poisson distribution with mean $\mu$ (the replay ratio constant). In other words, if $\mu=\frac{1}{2}$ then we replay $1$ old batch every $2$ new batches. This means that at the end of every new batch, $k$ batches are randomly sampled from the experience buffer, where $k$ follows the aforementioned Poisson distribution. During the sampling operation, we take a random number $z$ lower than the sum of all the priorities in the buffer, then the batch with the highest prefix sum lower than or equal to $z$ is sampled. \\
	Every time a batch is replayed, its priority is updated according to the new intrinsic rewards given by RND. \\
	Please, remember that intrinsic rewards are a measure of the novelty of a state, the more the agent explores the more some states may lose novelty. In other words, for the same state $s$ the intrinsic reward at time $t$ may significantly differ from the intrinsic reward at time $t+c$ with $c > 0$. \\
	Furthermore, similarly to \cite{horgan2018distributed}, all the aforementioned experience buffers are shared among all the workers of the A3C network.
	
	\subsection{Intrinsic Reward Replay}
	Intrinsic Rewards are given by the RND. The RND network is completely separated by the Actor and the Critic. The RND is continuously trained in order to properly evaluate the states in terms of their novelty: the less a state is seen by the agent, the more it is novel. This implies that high intrinsic rewards are associated to very uncommon states. For these reasons the Experience Replay with Intrinsic Rewards requires some precautions:
	\begin{enumerate}
		\item the RND can not be trained during the replay phase
		\item the intrinsic cumulative return of the replayed batches must be updated in order to properly compute the advantage and the critic loss
		\item the value used to get the advantage has always to be up to date
	\end{enumerate}
	Training the RND during replay would cause wrong intrinsic rewards, because the RND would start giving lower intrinsic rewards to the uncommon states, mostly because with ER we usually replay the uncommon states. 
	
	\section{Experiments} \label{sec:experiments}
	We conducted some experiments in order to prove the efficacy of PPO/RND extended with our Prioritized Oversampled Experience Replay technique. \\
	Mainly because our work is an extension of \cite{burda2018exploration} \footnote{and also because we do not have the hardware of OpenAI for testing on many more environments in less than a life-time}, we decided to focus our experiments primarily on the Atari game \textit{Montezuma's Revenge} and on a few other hard exploration games: \textit{Solaris} and \textit{Venture}. \\
	For PPO/RND\cite{burda2018exploration} we used the implementation publicly available at \cite{Github_PPORND}, while for the prioritized experience replay mechanism based on \cite{schaul2015prioritized} we used the code publicly available at \cite{Github_OpenAI_baselines} and then we quickly adapted it to our purposes.
	The changes we made to the default PPO/RND implementation are:
	\begin{itemize}
		\item We disabled the default intrinsic rewards scaling. In \cite{burda2018exploration} the intrinsic rewards are scaled by their running standard deviation.
		\item In the RND loss, we changed the dropout rate to $0.5$.
		\item We used A3C instead of A2C: we removed any synchronization barrier between the workers, thus updating the gradient using the Hogwild\cite{recht2011hogwild} approach.
		\item We used Proximal Value Optimization as in the default PPO\cite{schulman2017proximal} implementation.
		\item We optimized the vanilla A3C implementation in order to train more efficiently with GPUs. We did it through a technique called \quotes{delayed training} that is heavily inspired by \cite{espeholt2018impala}. This technique stores batches (even the replayed ones) into a buffer until a big-enough \textit{super-batch} is ready for training.
		\item We resized the games screen to a $42 \times 42$ (instead of $84 \times 84$ as in \cite{burda2018exploration}) grey-scaled image.
		\item We set the maximum number of allowed steps per episode to $3000$.
		\item We disabled sticky actions.
	\end{itemize}
	We think it is important to mention that, among the other things, we kept unchanged the following aspects of the default PPO/RND implementation:
	\begin{itemize}
		\item The CNN-based\footnote{with no RNNs} neural network.
		\item The extrinsic rewards clipping, in $[-1,1]$.
		\item The state representation, made of 4 consecutive grey-scaled screens.
	\end{itemize}
	The source of the code we used for our experiments is publicly available at \cite{Github_POER}. The aforementioned code is a new release of the code we published with \cite{amslaurea16718}, and it is meant to be easily extendible, readable and reusable.
	
	\subsection{Statistics for evaluation metrics} \label{sec:plotting-system}
	We have collected several statistics during training:
	\begin{itemize}
		\item \quotes{extrinsic\_reward}: average cumulative extrinsic reward per episode
		\item \quotes{extrinsic\_reward\_per\_step}: average cumulative extrinsic reward per step in an episode
		\item \quotes{extrinsic\_value\_per\_step}: average Critic extrinsic value per step in an episode
	\end{itemize}
	For every experiment we show how the \textit{mean} and \textit{standard deviation} of these statistics change during training time. Time is measured in \textit{parameters updates}, but in the last experiment (Section \ref{sec:exp4}) time is measured in \textit{steps} in order to focus more on sample efficiency.
	
	\subsection{Ablative Analysis on Montezuma's Revenge} \label{sec:analysis_montezuma}
	With our experiments we try to answer the following questions:
	\begin{enumerate}
		\item Is ER useful when combined with PPO/RND?
		\item Is experience prioritization useful in this context?
		\item Is prioritized drop useful in this context?
	\end{enumerate}
	We are going to show, through our ablative analysis, that extending PPO/RND with Prioritized Oversampled Experience Replay (POER) can improve sample efficiency without reducing exploration. \\
	In order to perform an ablative analysis we need to define a \textit{default} set of hyper-parameters, this way we are able to understand the impact of changing these parameters. \\
	In all the experiments, our \textit{default} set of hyper-parameters is the same described in \cite{burda2018exploration}, but due to the fact that we extended \cite{burda2018exploration} with POER, we have also the following extra hyper-parameters (see section \ref{sec:er_implementation} for more details):
	\begin{itemize}
		\item The replay frequency $\mu = 0.5$.
		\item The prioritized drop probability $P_d = 1$.
		\item The experience buffer size $B = 2^7$.
	\end{itemize}
	Furthermore, the super-batch size we adopted is $2^6$. This means that every super-batch is made of $2^6$ batches. \\
	In all the experiments, we used PPO trained on $t=128$ different Actor-Critics (ACs).
	
	\subsubsection{Is POER useful when combined with PPO/RND?} \label{sub:q1}
\begin{figure}[!h]
		\centering
		\includegraphics[width=0.5\textwidth]{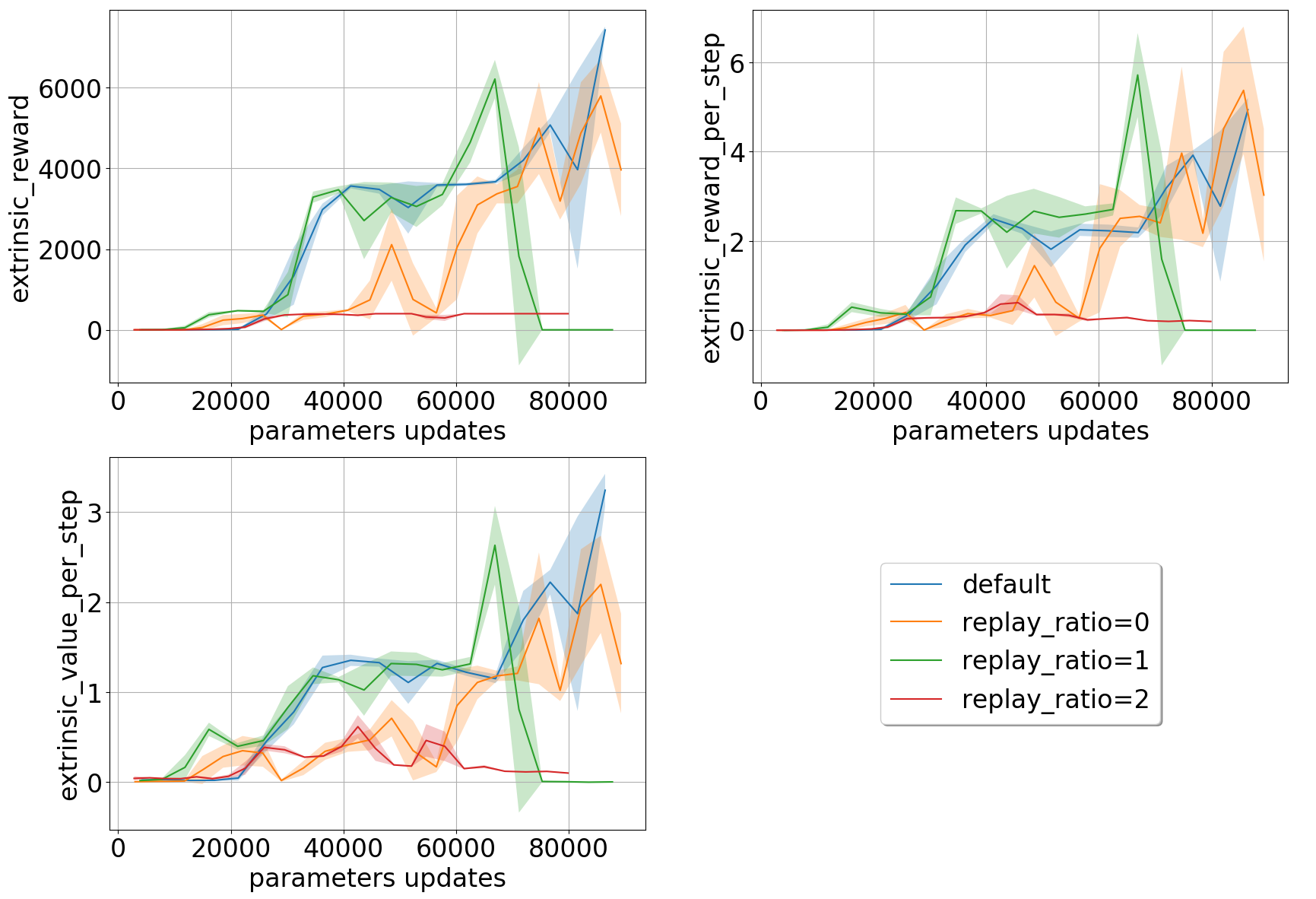}
		\caption{\textbf{Experiment 1 - Replay Frequency}}
		\label{fig:test1}
	\end{figure}

	With this experiment we try to understand the effects of PPO/RND combined with POER. We do it by changing the replay frequency $\mu$. We compare the \textit{default} experiment (having $\mu=0.5$) with 3 different experiments having respectively $\mu=0$, $\mu=1$ and $\mu=2$.
	The experiment having $\mu=0$ is said to be the \textit{baseline}, because it is equivalent to PPO/RND without any ER mechanism. \\
	In figure \ref{fig:test1} we show the training statistics. As you can see, during training the \textit{default} implementation of PPO/RND/POER (blue line) produces better results (in terms of mean extrinsic reward) than the \textit{baseline} (orange line). As expected, our implementation of PPO/RND combined with POER seems to be more sample efficient than the baseline, this is more evident when plotting the statistics against the \textit{steps} instead of the \textit{parameters updates} as shown in figure \ref{fig:test4} . \\
	It is interesting to notice that the replay frequency $\mu$ is a very important parameter to tune. In fact we can see that when $\mu$ is too big (eg. $\mu=2$), the exploratory skills of the agent tends to be worse. Intuitively this means that if we replay too much, then we also exploit too much the past experience without producing enough new information and thus losing in terms of exploration.
	
	\subsubsection{Is experience prioritization useful in this context?} \label{sub:q2}
\begin{figure}[!h]
		\centering
		\includegraphics[width=0.5\textwidth]{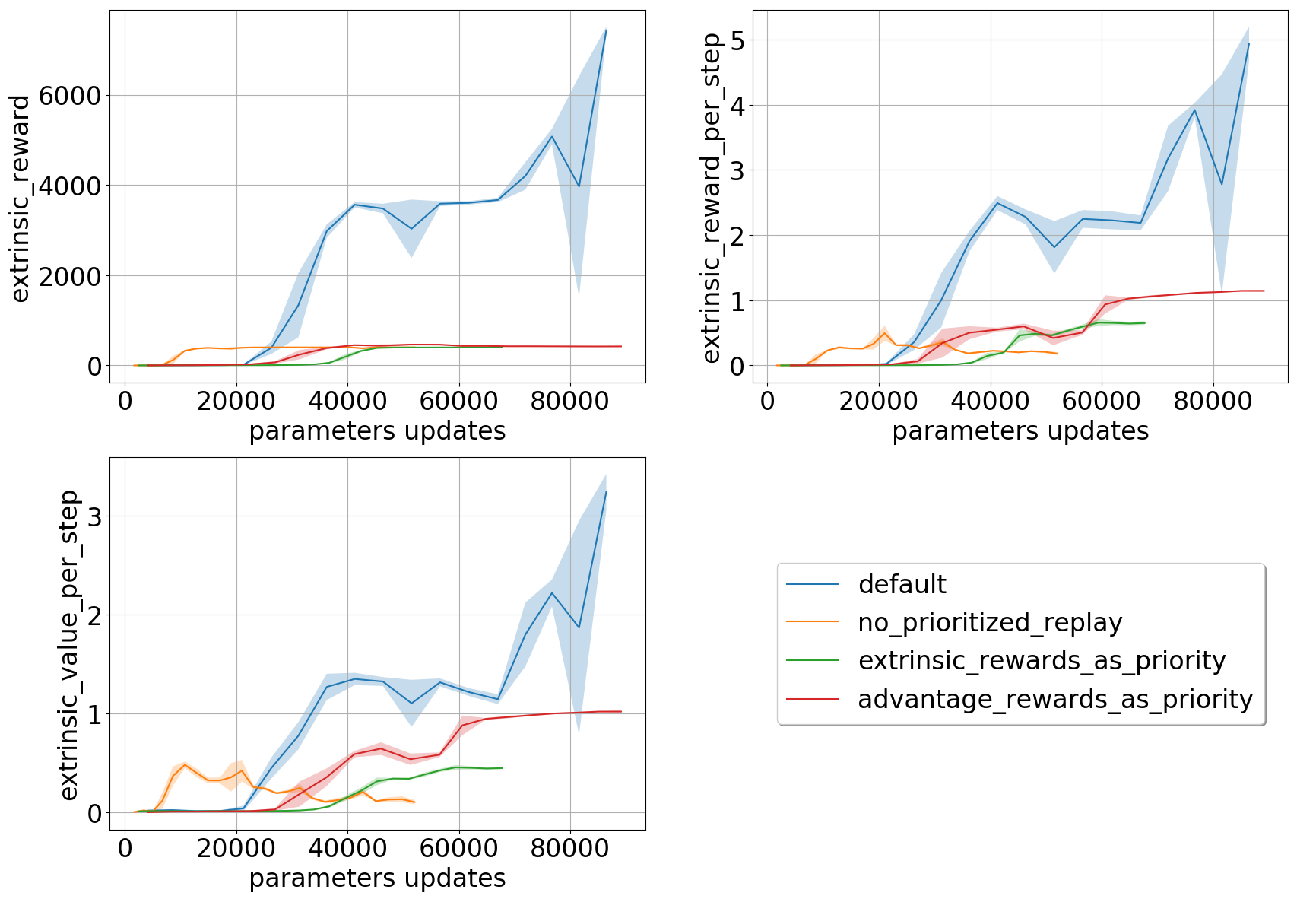}
		\caption{\textbf{Experiment 2 - Replay Prioritization}}
		\label{fig:test2}
	\end{figure}

	With this experiment we try to understand the effects of experience prioritization in PPO/RND combined with POER. We do it by disabling prioritization and by changing the priority function. We compare the \textit{default} (that uses \textit{ER prioritized with intrinsic rewards}) with three different experiments having respectively \textit{no prioritization}, \textit{ER prioritized with extrinsic rewards} and \textit{ER prioritized with advantages}. \\
	In figure \ref{fig:test2} we show the training statistics. As you can see, during training the \textit{default} (blue line) produces the best results, while the other experiments perform significantly worse. \\
	We believe that this fact supports the theory that we should mainly replay only \textit{uncommon batches} (identified by high intrinsic rewards).

	\subsubsection{Is prioritized drop useful in this context?} \label{sub:q3}
\begin{figure}[!h]
		\centering
		\includegraphics[width=0.5\textwidth]{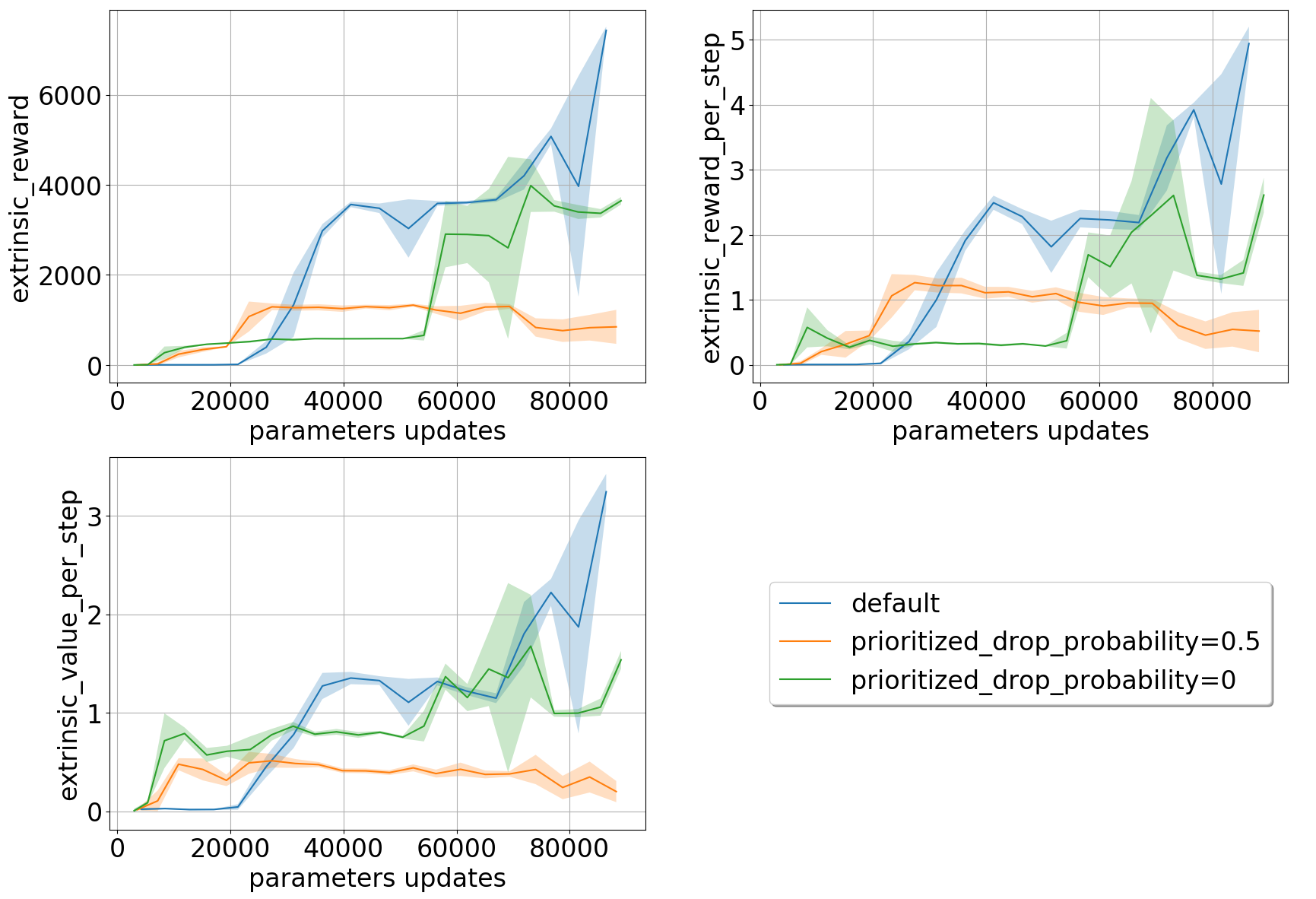}
		\caption{\textbf{Experiment 3 - Prioritized Drop Probability}}
		\label{fig:test3}
	\end{figure}

	With this experiment we try to understand the effects of changing the prioritized drop probability $P_d$ in PPO/RND combined with POER. We compare the \textit{default} (that has $P_d=1$) with two different experiments having respectively $P_d=0.5$ and $P_d=0$. \\
	In figure \ref{fig:test3} we show the training statistics. As you can see, during training the \textit{default} (blue line) produces the best results, while $P_d=0$ (green line) produces the worst results only initially. Thus, the prioritized drop seems an extremely important feature in POER. In other words, the strategy of replaying only the most uncommon important batches seems to be much more effective than randomly replaying both common and uncommon important batches.

	\subsection{Ablative Analysis on other hard exploration games} \label{sec:exp4}
\begin{figure}[!h]
		\centering
		\includegraphics[width=0.5\textwidth]{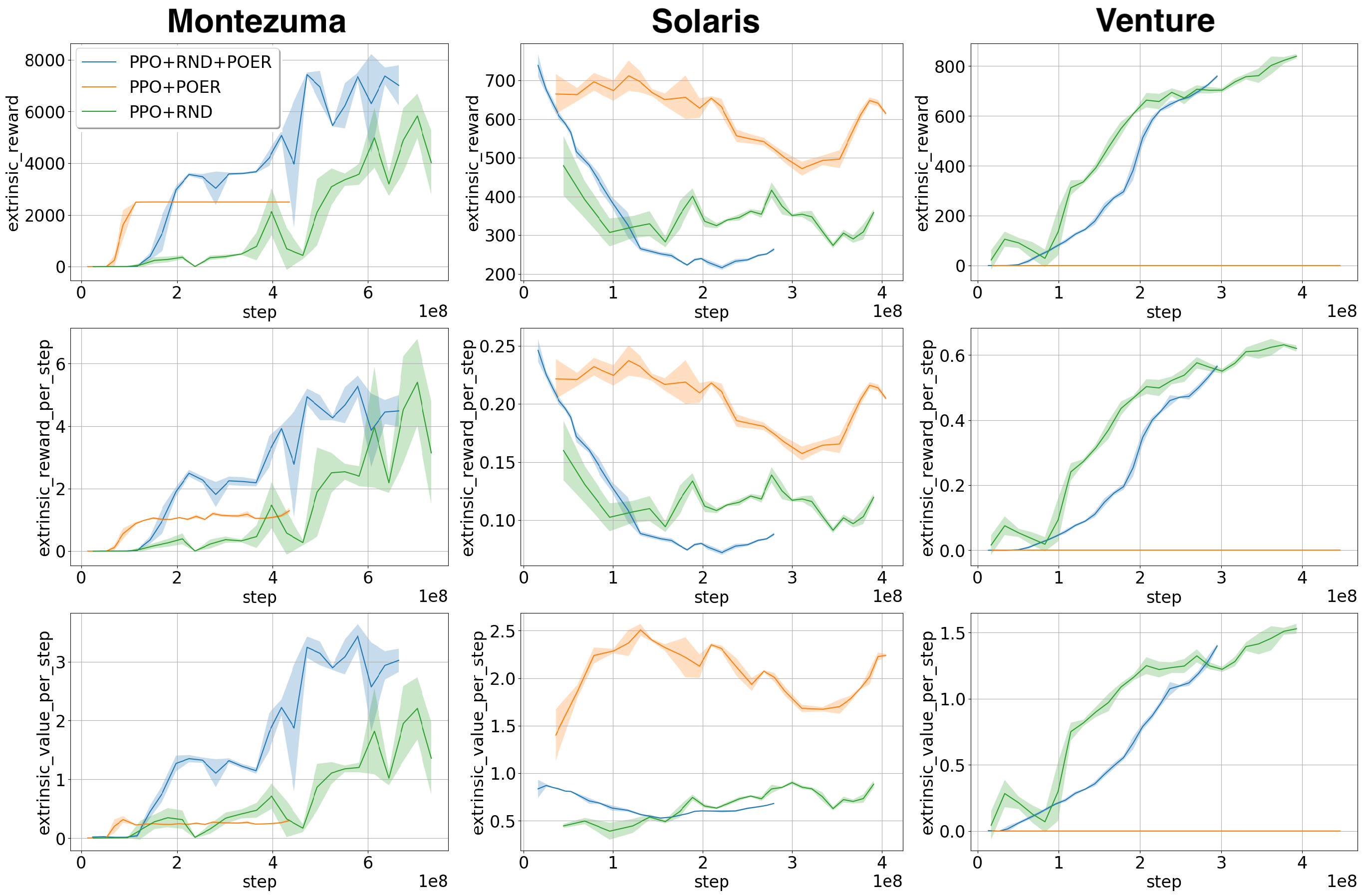}
		\caption{\textbf{Experiment 4 - Different Hard Exploration Games}}
		\label{fig:test4}
	\end{figure}

	In the previous ablative analysis we have shown that our combination of PPO/RND with POER, as expected, performs better in terms of sample efficiency than the baseline PPO/RND on the Atari game \textit{Montezuma's Revenge}. \\
	Technique-combination papers are somewhat rare in RL research because their results tend to have low significance: \textit{somehow we already expect independently good ideas for different aspects of an algorithm to combine somewhat well}. 
	Despite the aforementioned expectation, we will show with the following ablative analysis that our technique-combination does not always lead to performance improvements on hard exploration games, supporting the counter-intuitive fact that combining different ideas on different aspects of an algorithm does not always lead to better results. This is true especially when trying to combine techniques for exploration with techniques for exploitation. \\
	With this experiments we compare the performance of three different algorithms:
	\begin{itemize}
		\item PPO/RND/POER (using the \textit{default} set of hyper-parameters defined in section \ref{sec:analysis_montezuma}).
		\item PPO/RND (the \textit{default} but with $\mu = 0$).
		\item PPO/POER (the \textit{default} but with no intrinsic rewards and thus with ER prioritized by extrinsic rewards)
	\end{itemize}
	in 3 different hard exploration \cite{bellemare2016unifying} Atari games:
	\begin{itemize}
		\item \textit{Montezuma's Revenge}: characterized by very sparse rewards, also difficult to get by random actions because the agent can die very easily.
		\item \textit{Solaris}: compared to \textit{Montezuma} it is characterized by more frequent rewards that are also easier to get by random actions.
		\item \textit{Venture}: characterized by sparse rewards but apparently easier to get by random actions than \textit{Montezuma}.
	\end{itemize}
	In order to focus more on sample efficiency, in figure \ref{fig:test4} we show the training statistics plotted against the \textit{steps} instead of the \textit{parameters updates}. As we can see: in \textit{Montezuma's Revenge} the best algorithm is PPO/RND/POER (blue line), in \textit{Solaris} the best one is PPO/POER (orange line), while in \textit{Venture} only those algorithms using RND perform decently enough and, as expected, PPO/RND/POER is the most sample efficient. In all the aforementioned games we can see that the statistics of the algorithms using POER seem to have a lower \textit{standard deviation} and a more stable \textit{mean} than PPO/RND, and we think this is an indicator of the better sample efficiency of PPO/RND/POER with respect to PPO/RND. \\
	We believe that tuning further the replay frequency $\mu$ might help to improve the performance of PPO/RND/POER in both \textit{Solaris} and \textit{Venture}. 

	\section{CONCLUSIONS}
	In this paper we have shown a simple way to combine Experience Replay with Intrinsic Rewards, or in other words a simple way to balance exploration with exploitation. \\
	In order to do that, we:
	\begin{itemize}
		\item give a definition of \textit{important} batch
		\item find a way to know when a batch is \textit{uncommon}
		\item understand how to use important uncommon batches to improve the exploratory skills of the agent
	\end{itemize}
	Furthermore, through our experiments we try to answer the following questions:
	\begin{enumerate}
		\item Is ER useful when combined with PPO/RND?
		\item Is experience prioritization useful in this context?
		\item Is prioritized drop useful in this context?
	\end{enumerate}
	The answers we got are:
	\begin{enumerate}
		\item ER seems very useful, but a too high replay frequency might unbalance the agent toward the exploitation of old information, losing in terms of exploratory skills, especially in environments characterized by more frequent extrinsic rewards than \textit{Montezuma}. 
		\item Experience prioritization based on intrinsic rewards is definitively important in order to replay the uncommon batches.
		\item A fully prioritized drop seems to give the best performances. The strategy of replaying only the most uncommon important batches seems to be much more effective than randomly replaying both common and uncommon important batches.
	\end{enumerate}

	\bibliographystyle{IEEEtran}
	\bibliography{biblio.bib}

\begin{thebibliography}{10}
\providecommand{\url}[1]{#1}
\csname url@samestyle\endcsname
\providecommand{\newblock}{\relax}
\providecommand{\bibinfo}[2]{#2}
\providecommand{\BIBentrySTDinterwordspacing}{\spaceskip=0pt\relax}
\providecommand{\BIBentryALTinterwordstretchfactor}{4}
\providecommand{\BIBentryALTinterwordspacing}{\spaceskip=\fontdimen2\font plus
\BIBentryALTinterwordstretchfactor\fontdimen3\font minus
  \fontdimen4\font\relax}
\providecommand{\BIBforeignlanguage}[2]{{%
\expandafter\ifx\csname l@#1\endcsname\relax
\typeout{** WARNING: IEEEtran.bst: No hyphenation pattern has been}%
\typeout{** loaded for the language `#1'. Using the pattern for}%
\typeout{** the default language instead.}%
\else
\language=\csname l@#1\endcsname
\fi
#2}}
\providecommand{\BIBdecl}{\relax}
\BIBdecl

\bibitem{burda2018exploration}
Y.~Burda, H.~Edwards, A.~Storkey, and O.~Klimov, ``Exploration by random
  network distillation,'' \emph{arXiv preprint arXiv:1810.12894}, 2018.

\bibitem{schulman2017proximal}
J.~Schulman, F.~Wolski, P.~Dhariwal, A.~Radford, and O.~Klimov, ``Proximal
  policy optimization algorithms,'' \emph{arXiv preprint arXiv:1707.06347},
  2017.

\bibitem{jaderberg2016reinforcement}
M.~Jaderberg, V.~Mnih, W.~M. Czarnecki, T.~Schaul, J.~Z. Leibo, D.~Silver, and
  K.~Kavukcuoglu, ``Reinforcement learning with unsupervised auxiliary tasks,''
  \emph{arXiv preprint arXiv:1611.05397}, 2016.

\bibitem{schaul2015prioritized}
T.~Schaul, J.~Quan, I.~Antonoglou, and D.~Silver, ``Prioritized experience
  replay,'' \emph{arXiv preprint arXiv:1511.05952}, 2015.

\bibitem{bellemare2016unifying}
M.~Bellemare, S.~Srinivasan, G.~Ostrovski, T.~Schaul, D.~Saxton, and R.~Munos,
  ``Unifying count-based exploration and intrinsic motivation,'' in
  \emph{Advances in Neural Information Processing Systems}, 2016, pp.
  1471--1479.

\bibitem{wang2016sample}
Z.~Wang, V.~Bapst, N.~Heess, V.~Mnih, R.~Munos, K.~Kavukcuoglu, and
  N.~de~Freitas, ``Sample efficient actor-critic with experience replay,''
  \emph{arXiv preprint arXiv:1611.01224}, 2016.

\bibitem{oh2018self}
J.~Oh, Y.~Guo, S.~Singh, and H.~Lee, ``Self-imitation learning,'' \emph{arXiv
  preprint arXiv:1806.05635}, 2018.

\bibitem{yang2019never}
H.-K. Yang, P.-H. Chiang, K.-W. Ho, M.-F. Hong, and C.-Y. Lee, ``Never forget:
  Balancing exploration and exploitation via learning optical flow,''
  \emph{arXiv preprint arXiv:1901.08486}, 2019.

\bibitem{amslaurea16718}
\BIBentryALTinterwordspacing
F.~Sovrano, ``Deep reinforcement learning and sub-problem decomposition using
  hierarchical architectures in partially observable environments,'' Master's
  thesis, Università di Bologna, 2018. [Online]. Available:
  \url{http://amslaurea.unibo.it/16718/}
\BIBentrySTDinterwordspacing

\bibitem{asperti2019crawling}
A.~Asperti, D.~Cortesi, C.~De~Pieri, G.~Pedrini, and F.~Sovrano, ``Crawling in
  rogue's dungeons with deep reinforcement techniques,'' \emph{IEEE
  Transactions on Games}, 2019.

\bibitem{asperti2018crawling}
A.~Asperti, D.~Cortesi, and F.~Sovrano, ``Crawling in rogue's dungeons with
  (partitioned) a3c,'' \emph{arXiv preprint arXiv:1804.08685}, 2018.

\bibitem{mnih2013playing}
V.~Mnih, K.~Kavukcuoglu, D.~Silver, A.~Graves, I.~Antonoglou, D.~Wierstra, and
  M.~Riedmiller, ``Playing atari with deep reinforcement learning,''
  \emph{arXiv preprint arXiv:1312.5602}, 2013.

\bibitem{L2Loss}
Tensorflow, ``Tensorflow's l2 loss,''
  \url{https://www.tensorflow.org/api_docs/python/tf/nn/l2_loss}.

\bibitem{mnih2016asynchronous}
V.~Mnih, A.~P. Badia, M.~Mirza, A.~Graves, T.~Lillicrap, T.~Harley, D.~Silver,
  and K.~Kavukcuoglu, ``Asynchronous methods for deep reinforcement learning,''
  in \emph{International Conference on Machine Learning}, 2016, pp. 1928--1937.

\bibitem{stooke2018accelerated}
A.~Stooke and P.~Abbeel, ``Accelerated methods for deep reinforcement
  learning,'' \emph{arXiv preprint arXiv:1803.02811}, 2018.

\bibitem{schulman2015trust}
J.~Schulman, S.~Levine, P.~Abbeel, M.~Jordan, and P.~Moritz, ``Trust region
  policy optimization,'' in \emph{International Conference on Machine
  Learning}, 2015, pp. 1889--1897.

\bibitem{lin1992memory}
L.-J. Lin and T.~M. Mitchell, \emph{Memory approaches to reinforcement learning
  in non-Markovian domains}.\hskip 1em plus 0.5em minus 0.4em\relax Citeseer,
  1992.

\bibitem{Qlearning15}
\BIBentryALTinterwordspacing
V.~Mnih, K.~Kavukcuoglu, D.~Silver, A.~A. Rusu, J.~Veness, M.~G. Bellemare,
  A.~Graves, M.~A. Riedmiller, A.~Fidjeland, G.~Ostrovski, S.~Petersen,
  C.~Beattie, A.~Sadik, I.~Antonoglou, H.~King, D.~Kumaran, D.~Wierstra,
  S.~Legg, and D.~Hassabis, ``Human-level control through deep reinforcement
  learning,'' \emph{Nature}, vol. 518, no. 7540, pp. 529--533, 2015. [Online].
  Available: \url{https://doi.org/10.1038/nature14236}
\BIBentrySTDinterwordspacing

\bibitem{pathak2017curiosity}
D.~Pathak, P.~Agrawal, A.~A. Efros, and T.~Darrell, ``Curiosity-driven
  exploration by self-supervised prediction,'' in \emph{International
  Conference on Machine Learning (ICML)}, vol. 2017, 2017.

\bibitem{ostrovski2017count}
G.~Ostrovski, M.~G. Bellemare, A.~van~den Oord, and R.~Munos, ``Count-based
  exploration with neural density models,'' in \emph{Proceedings of the 34th
  International Conference on Machine Learning-Volume 70}.\hskip 1em plus 0.5em
  minus 0.4em\relax JMLR. org, 2017, pp. 2721--2730.

\bibitem{tang2017exploration}
H.~Tang, R.~Houthooft, D.~Foote, A.~Stooke, O.~X. Chen, Y.~Duan, J.~Schulman,
  F.~DeTurck, and P.~Abbeel, ``\# exploration: A study of count-based
  exploration for deep reinforcement learning,'' in \emph{Advances in Neural
  Information Processing Systems}, 2017, pp. 2753--2762.

\bibitem{chentanez2005intrinsically}
N.~Chentanez, A.~G. Barto, and S.~P. Singh, ``Intrinsically motivated
  reinforcement learning,'' in \emph{Advances in neural information processing
  systems}, 2005, pp. 1281--1288.

\bibitem{schmidhuber1991possibility}
J.~Schmidhuber, ``A possibility for implementing curiosity and boredom in
  model-building neural controllers,'' in \emph{Proc. of the international
  conference on simulation of adaptive behavior: From animals to animats},
  1991, pp. 222--227.

\bibitem{still2012information}
S.~Still and D.~Precup, ``An information-theoretic approach to curiosity-driven
  reinforcement learning,'' \emph{Theory in Biosciences}, vol. 131, no.~3, pp.
  139--148, 2012.

\bibitem{csimcsek2006intrinsic}
{\"O}.~{\c{S}}im{\c{s}}ek and A.~G. Barto, ``An intrinsic reward mechanism for
  efficient exploration,'' in \emph{Proceedings of the 23rd international
  conference on Machine learning}.\hskip 1em plus 0.5em minus 0.4em\relax ACM,
  2006, pp. 833--840.

\bibitem{horgan2018distributed}
D.~Horgan, J.~Quan, D.~Budden, G.~Barth-Maron, M.~Hessel, H.~Van~Hasselt, and
  D.~Silver, ``Distributed prioritized experience replay,'' \emph{arXiv
  preprint arXiv:1803.00933}, 2018.

\bibitem{Github_PPORND}
Y.~Burda, H.~Edwards, A.~Storkey, and O.~Klimov, ``Exploration by random
  network distillation,''
  \url{https://github.com/openai/random-network-distillation}, 2018.

\bibitem{Github_OpenAI_baselines}
P.~Dhariwal, C.~Hesse, O.~Klimov, A.~Nichol, M.~Plappert, A.~Radford,
  J.~Schulman, S.~Sidor, Y.~Wu, and P.~Zhokhov, ``Openai baselines,''
  \url{https://github.com/openai/baselines}, 2017.

\bibitem{recht2011hogwild}
B.~Recht, C.~Re, S.~Wright, and F.~Niu, ``Hogwild: A lock-free approach to
  parallelizing stochastic gradient descent,'' in \emph{Advances in neural
  information processing systems}, 2011, pp. 693--701.

\bibitem{espeholt2018impala}
L.~Espeholt, H.~Soyer, R.~Munos, K.~Simonyan, V.~Mnih, T.~Ward, Y.~Doron,
  V.~Firoiu, T.~Harley, I.~Dunning \emph{et~al.}, ``Impala: Scalable
  distributed deep-rl with importance weighted actor-learner architectures,''
  \emph{arXiv preprint arXiv:1802.01561}, 2018.

\bibitem{Github_POER}
F.~Sovrano, ``Combining 'experience replay' with 'exploration by random network
  distillation',''
  \url{https://github.com/Francesco-Sovrano/Combining--experience-replay--with--exploration-by-random-network-distillation-},
  2019.

\end{thebibliography}
	
\end{document}